# Label Information Enhanced Fraud Detection against Low Homophily in Graphs

Yuchen Wang*
Southeast University
Nanjing, China
yuchen_seu@seu.edu.cn

Jinghui Zhang†
Southeast University
Nanjing, China
jhzhang@seu.edu.cn

Zhengjie Huang
Baidu Inc.
Shenzhen, China
huangzhengjie@baidu.com

Weibin Li
Baidu Inc.
Shenzhen, China
liweibin02@baidu.com

Shikun Feng
Baidu Inc.
Shenzhen, China
fengshikun01@baidu.com

Ziheng Ma
Baidu Inc.
Beijing, China
maziheng01@baidu.com

Yu Sun
Baidu Inc.
Beijing, China
sunyu02@baidu.com

Dianhai Yu
Baidu Inc.
Beijing, China
yudianhai@baidu.com

Fang Dong
Southeast University
Nanjing, China
fdong@seu.edu.cn

Jiahui Jin
Southeast University
Nanjing, China
jjin@seu.edu.cn

Beilun Wang
Southeast University
Nanjing, China
beilun@seu.edu.cn

Junzhou Luo
Southeast University
Nanjing, China
jluo@seu.edu.cn

## ABSTRACT

Node classification is a substantial problem in graph-based fraud detection. Many existing works adopt Graph Neural Networks (GNNs) to enhance fraud detectors. While promising, currently most GNN-based fraud detectors fail to generalize to the low homophily setting. Besides, label utilization has been proved to be significant factor for node classification problem. But we find they are less effective in fraud detection tasks due to the low homophily in graphs. In this work, we propose GAGA, a novel **G**roup **AG**gregation enhanced Tr**A**nsformer, to tackle the above challenges. Specifically, the group aggregation provides a portable method to cope with the low homophily issue. Such an aggregation explicitly integrates the label information to generate distinguishable neighborhood information. Along with group aggregation, an attempt towards end-to-end trainable group encoding is proposed which augments the original feature space with the class labels. Meanwhile, we devise two additional learnable encodings to recognize the structural and relational context. Then, we combine the group aggregation and the learnable encodings into a Transformer encoder to capture the semantic information. Experimental results clearly show that GAGA outperforms other competitive graph-based fraud detectors by up to 24.39% on two trending public datasets and a real-world industrial dataset from Baidu. Even more, the group aggregation is demonstrated to outperform other label utilization methods (e.g., C&S, BoT/UniMP) in the low homophily setting.

---

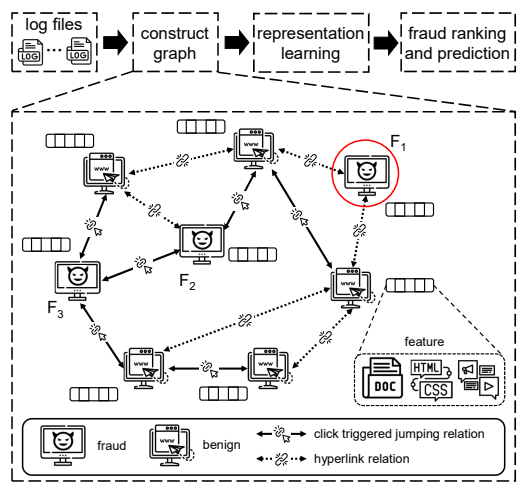

Figure 1: A toy multi-relational fraud graph of a website anti-fraud task. The target node $F_1$ highlighted with a colored circle is a fraudulent website that disguises itself by connecting to benign nodes (i.e., low homophily patterns).

## 1 INTRODUCTION

With the expansion of the Internet, fraud has become a pernicious problem involving various kinds of fraudulent entities, e.g., deceptive reviews, fake online accounts and malicious websites. One of the most frequently applied fraud detection tasks is entity classification. In recent years, graph-based fraud detection approaches have been broadly applied in many practical applications. By modeling the entities as (attributed) nodes and the corresponding interactions as edges, the suspicious entities can be detected at the graph level. For example, one might wish to classify the deceptive reviews in a *Review-User-Review* graph [4], predict the fake accounts in a *User-IP-User* network [30], or filter the malicious websites with the *Click Triggered Jumping* interactions [33].

However, many existing graph-based approaches overlook the adversarial actions of fraudulent entities discussed in [5, 8, 9]. As shown in Fig. 1, to illustrate the issue, we construct a multi-relational fraud graph using the dataset collected by Baidu. We first transform the log files into a fraud graph structure, where the nodes correspond to the unique domains and the edges to the multiple relations (i.e., click triggered jumping relation and hyperlink relation, respectively) between domains. Each node also consists of a feature vector extracted by a toolkit [23] to describe the textual information. In practical applications, the adversarial actions of fraudulent entities might induce *heterophily or low homophily* [34] where the target node (e.g., node $F_1$ in Fig. 1) and its immediate neighbors are prone to have different class labels and dissimilar features. It is also noteworthy that in many scenarios, entities could have high-order interactions (e.g., fraud node $F_1$ and $F_2$ are two-hop neighbors), or be connected with multiple types of relations (e.g., $F_2$ connects to its neighbors via two types of edges).

Given such a multi-relational fraud graph, a typical solution is to design a graph-based algorithm to exploit the suspicious patterns and classify fraudulent entities. Recent years have witnessed research efforts devoting to deep learning methods on graphs. GNNs are crucial tools for graph representation learning and proved to be effective over massive tasks [29]. Built upon message passing, GNNs update the representation of the target node by recursively aggregating neighborhood information with neural modules.

**Challenges.** While the GNN-based fraud detectors seem promising, two essential challenges still remain to be further explored. *Challenge 1: The low homophily issue impairs the performance of GNN-based approaches.* Through the aggregation, the representation of the target node is assimilated by its neighbor's representation regardless of whether they share the same label. Such an indiscriminate aggregation is not suitable for graph-based fraud detection and will deteriorate the performance of GNN-based detectors. *Challenge 2: The limited capability of existing approaches in taking full advantage of label information.* GNN-based fraud detectors leverage a parameterized message passing mechanism to learn predictive representations from the features of neighboring nodes. These approaches are confined to the feature space and only consider the node labels as the supervision signal in the training phase. Therefore, the potential usefulness of the label information in augmenting original node features are overlooked in previous work which would impact the detection performance of fraud detectors.

**Current attempts.** CARE-GNN [4] and PC-GNN [14] focus on neighbor filtering to mitigate the low homophily issue. Both approaches select immediate neighbors with high similarity scores for aggregation. $H^2$-FDetector [21] uses the labeled node as supervision to identify homophilic and heterophilic connections and performs separate aggregations. Nonetheless, these enhanced GNN-based fraud detectors does not take full advantage of the label information. The label propagation algorithms [7, 35] smooth labels across the graph while the unified message passing [22, 27] considers a random subset of labels, along with the feature, as the input of GNNs. Unfortunately, while effective in many node classification scenarios (e.g., Open Graph Benchmark (OGB) leaderboard [13]), these label utilization methods do not always take effect, especially in the binary classification and low homophily setting [22].

**Our contributions.** In an effort to devise a label information enhanced fraud detector, we first introduce a simple yet effective preprocessing strategy named Group Aggregation (GA). Then the original neighborhood features are grouped into sequential data. And secondly, we provide a learnable encoding method that retains the context of *structural*, *relational* and *label* information. Along with the multi-head self-attention mechanism of the backbone encoder, we can capture the preference of grouped features. The contributions are summarized as follows:

- **Group Aggregation:** GA first overcomes *Challenge 1* based on the fact that the class labels indicate the homophily traits. Instead of aggregating all the neighborhood information into single vector, GA explicitly uses the class labels to distinguish the neighborhood information. For each target node, GA can be generalized to the high-order neighbors with respect to multiple relations.
- **Learnable Encodings:** To solve *Challenge 2*, we propose a novel *Group Encoding* to enrich the original feature space with class label augmented features at both training and inference time. Additionally, we devise a *Hop Encoding* and a *Relation Encoding* to reflect the multi-hop adjacency information and the order of relations in the grouped vectors generated by GA. The learnable encodings can be integrated into the Transformer encoder and automatically capture information in an end-to-end fashion.
- **Effectiveness:** The proposed GAGA achieves up to 24.39% performance improvement over previous state-of-the-art fraud detectors on two public datasets and a real-world industrial dataset.

## 2 RELATED WORK

**Advances in node-level classification.** Semi-supervised node classification aims to learn from the labeled data and predict the properties of the unlabeled nodes [10]. DeeperGCN [12] introduces residual connections and message normalization layers to make GCNs go deeper. SIGN [20] decouples feature aggregation from the training time to scale to large graphs. Besides, sampling based methods, such as GraphSAGE [6], Cluster-GCN [3] and GraphSAINT [31], build minibatches from subgraphs for large-scale graph representation learning. Regarding label utilization, Label Propagation Algorithm (LPA) [35] propagates labels via edges to make predictions. Correct and Smooth (C&S) [7] provides a post-processing method that utilizes labels to help boost prediction performance. In particular, UniMP [22] combines node feature propagation with label embeddings at both training and inference time. Apart from the above approaches, there has been a growing interest in Transformer-based graph representation learning [2, 21, 22]. Inspired by the Transformer architecture [24], some previous attempts replace the aggregator with the self-attention module, but the limitations of classic message passing hinder the applicability of these models.

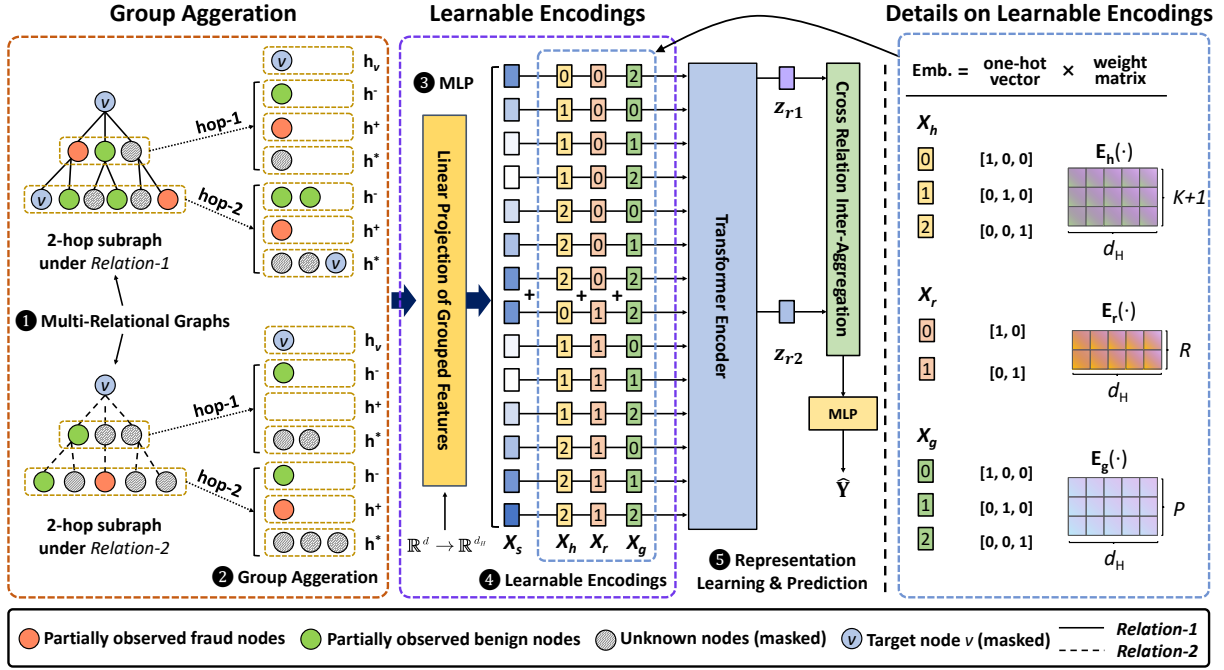

Figure 2: The overall architecture of GAGA. The number of hops $K$ and relations $R$ are set to 2 for simplicity.

**Graph-based fraud detection.** Recently, many graph-based fraud detectors have been proposed [8, 11, 16, 17]. CARE-GNN [4] and RioGNN [19] are GNN-based fraud detectors with reinforced neighbor selection module, which concentrates on tackling the camouflage problem in fraud graphs. PC-GNN [14] introduces a node sampler and a label-aware neighbor selector to further re-weight imbalanced classes. FRAUDRE [32] devises fraud-aware graph convolution module and imbalance-oriented optimization objective dual-resistant to graph inconsistency and imbalance. $H^2$-FDetector [21] identifies the homophilic and heterophilic connections and applies separate aggregation strategies separately for different types of connections, which is the most similar work. TADDY [15] proposes a Transformer-based detector for dynamic graphs and constructs additional encodings to capture structural and temporal patterns. However, these GNN-based methods suffer from the inherent limitations of original message passing. In this paper, we attempt to tackle the above challenges with the group aggregation and learnable encodings.

## 3 PRELIMINARIES

In this section, we formulate the graph-based fraud detection as a semi-supervised node-level binary classification problem. We first define a general procedure as follows.

**Multi-relational fraud graph construction.** The entities, relations and features can be modeled as a multi-relational fraud graph $\mathcal{G}(\mathcal{V}, \mathcal{E}, \mathcal{X}, \mathcal{Y})$, that consists of a node set of $N$ entities $\mathcal{V} = \{v_1, v_2, ..., v_N\}$ ($N = |\mathcal{V}|$), $R$ adjacency matrices of corresponding relations $\mathcal{E} = \{\mathbf{A}_1, \mathbf{A}_2, ..., \mathbf{A}_R\}$ ($R = |\mathcal{E}|$), a set of node feature vectors $\mathcal{X} = \{\mathbf{x}_1, \mathbf{x}_2, ..., \mathbf{x}_N\}$ and a node label set $\mathcal{Y}$. Nodes $u$ and $v$ are connected under relation $r \in \{1, 2, ..., R\}$ if $\mathbf{A}_r[u, v] = 1$. Each node $v \in \mathcal{V}$ has a $d$-dimension feature vector $\mathbf{x}_v \in \mathbb{R}^d$. In the graph-based fraud detection, we consider a semi-supervised situation where only a small portion of nodes (denoted by $\hat{\mathcal{V}}$) in $\mathcal{G}$ are labeled (denoted by $\hat{\mathcal{Y}}$). Each node $v \in \hat{\mathcal{V}}$ has a binary scalar label $y_v \in \{0, 1\}$, where $y = 1$ represents fraud nodes and $y = 0$ represents benign nodes. Hence, the number of classes in fraud graphs is defined as $C = 2$.

**Representation learning and fraud detection.** Given graph $\mathcal{G}$, the detection process is to learn discriminative embedding and output class probabilities for each node based on the node feature set $\mathcal{X}$, the multi-perspective relations $\mathcal{E}$ and the partially observed labels $\hat{\mathcal{Y}}$. In practical, a predefined threshold is usually given to determine the labeling of target nodes, and thus nodes that have above-threshold probabilities are identified as fraud entities.

## 4 THE PROPOSED APPROACH

### 4.1 Model Overview

As illustrated in Fig. 2, GAGA consists mainly of group aggregation (❶, ❷), learnable encodings (❸, ❹) and representation learning and prediction (❺). In the group aggregation step, we first precompute the multi-hop neighborhood information of each target node and perform group aggregation to generate a sequence of group vectors as the input. Next, in the learnable encodings step, the sequence of group vectors are encoded with three types of learnable embeddings to retain structural, relational and label information. We conduct multi-head self-attention to adaptively re-weights these group vectors, which can capture more similar information under low homophily settings. At last, we combine the output vectors across different relations and generate the final representation of the target node. A following additional Multi-Layer Perceptron neural network (MLP) serves as the prediction head.

## 4.2 Group Aggregation

The previous neighborhood aggregation function can be formulated as follows (assuming self-loop updated):

$$f_{agg}(\{\mathbf{x}_u | \forall u \in \mathcal{N}(v) \cup \{v\}\}) = \frac{1}{\phi(\cdot)} \sum_{u \in \mathcal{N}(v) \cup \{v\}} \mathbf{x}_u, \quad (1)$$

where $\phi(\cdot)$ is a differentiable normalization function, $\mathcal{N}(v)$ is the immediate neighbors of node $v$. However, such an aggregation fails to generalize to the low homophily setting.

*4.2.1 One-hop group aggregation.* Motivated by the above limitation, we devise a group aggregation strategy to highlight the information of different classes. The key idea lies in incorporating partially observed labels into the aggregation process and re-weighting neighbor information of different classes. As distinguished from the single-output aggregation in Equation 1, group aggregation provides multiple outputs. Nodes with the same class label come into the same group and then each group performs aggregation separately. Since the label is explicitly used to group neighbors, the target node may be also in the neighbor set via self-loop or multi-hop propagation. A preprocessing step named *target node masking* is implemented to avoid label leakage.

Concretely, for a target node $v$ and its one-hop neighbors with self-loop $\hat{\mathcal{N}}_1(v) = \mathcal{N}(v) \cup \{v\}$ under relation $r$, the nodes are divided into $P$ groups $V = \{V_1, V_2, \ldots, V_C, V^*\}$ ($P = C + 1$, the last group $V_P$ is defined as $V^*$ since we put masked nodes in $V_P$). Therefore, we can perform the group aggregation and then get a sequence of $P$ group vectors:

$$\begin{aligned}\mathbf{H}_g &= [\mathbf{h}_1, \mathbf{h}_2, \ldots, \mathbf{h}_C, \mathbf{h}^*] \\ \mathbf{h}_i &= f_{agg}(\{\mathbf{x}_u | \forall u \in V_i\}) = \frac{1}{\phi(\cdot)} \sum_{u \in V_i} \mathbf{x}_u,\end{aligned} \quad (2)$$

where $\mathbf{H}_g$ is the sequence of group vectors, the normalization function is defined as $\phi(\cdot) = |V_i|^\alpha$ ($\alpha$ is a constant), and $\mathbf{h}^*$ is the aggregation result of masked nodes whose labels are unknown. In the multi-relation fraud graph, the class label is binary, so the information of the target node and its one-hop neighbors can be aggregated individually into three group vectors $\mathbf{H}_g = [\mathbf{h}_1, \mathbf{h}_2, \mathbf{h}_3] = [\mathbf{h}^-, \mathbf{h}^+, \mathbf{h}^*]$, namely benign group $V^-$ of negative labels (−), fraud group $V^+$ of positive labels (+) and unknown group $V^*$ of masked labels (∗), respectively. For those empty groups, we patch zeros as default.

*4.2.2 Multi-hop group aggregation.* To access distant neighbors and capture high-order information, we generalize the group aggregation to multi-hop neighborhood. First, for each target node $v$, we precompute the full $K$-hop neighbors $\{\hat{\mathcal{N}}_i(v) | i = 1, 2, \ldots, K\}$, and apply group aggregation for each hop's nodes (Fig. 2 ❷):

$$\begin{aligned}\mathbf{H}_r &= \|_{k=1}^{K} \mathbf{H}_g^{(k)}, \\ \mathbf{H}_g^{(k)} &= [\mathbf{h}^-, \mathbf{h}^+, \mathbf{h}^*]^{(k)} \text{ given } \hat{\mathcal{N}}_k(v),\end{aligned} \quad (3)$$

where $\mathbf{H}_r$ is the group aggregation results of neighborhood information within $K$ hops under relation $r$, $\mathbf{H}_g^{(k)}$ represents the group vectors of the $k$-th hop, and $\|$ denotes concatenation operation. In order to apply self-attention on the target node and perform classification, we combine the raw feature $\mathbf{h}_v$ and the group vectors $\mathbf{H}_r$ together into single sequence $\mathbf{H}_{v,r} = [\mathbf{h}_v] \| \mathbf{H}_r$. After performing group aggregation under each relation, the next step is to combine all the sequences $\mathbf{H}_{v,r}$ as the input feature sequence, which is defined as $\mathbf{H}_s = \|_{r=1}^{R} \mathbf{H}_{v,r}$. As illustrated in Fig. 2 ❹, the amount of vectors in the input feature sequence is $S = R \times (P \times K + 1)$, where $S$ is the sequence length.

*4.2.3 Remark.* In the inductive anti-fraud scenario, we can perform GA for fresh individual nodes, which is similar to the sampling process of GraphSAGE [6]. For those isolated entities that are extremely deficiency of labeled neighbors, most nodes would belong to the unknown group. Such a situation reduces GA to normal aggregation in GNNs. We perform ablation studies and discuss the issue in Experiments and Appendix.

## 4.3 Learnable Encodings for Graph Transformer

GA models the high-dimensional graph structure into sequential data. To exploit the structural, relational and label information in the original multi-relation fraud graphs, three learnable encodings are introduced (Fig. 2 ❹). Meanwhile, we propose to directly adopt a transformer-based encoder to learn representations from the input feature sequences. By means of the multi-head self-attention, the vectors from different hops, relations and groups are re-weighted automatically.

*4.3.1 Linear projection layer.* To generate the final input sequence of the transformer encoder, we first linearly embed each of the group vectors $\mathbf{X}_s = \sigma(\psi(\mathbf{H}_s))$, where function $\psi : \mathbb{R}^{S \times d} \to \mathbb{R}^{S \times d_H}$, $d$ and $d_H$ represents the dimension of encoding, and $\sigma$ is the activation function (Fig. 2 ❸).

*4.3.2 Hop encoding.* The transformer-based encoder independently treats each group vector without distinction. To make the encoder aware of the multi-hop structural information, the hop encoding is necessary. Given a target node $v$ and its feature sequence $\mathbf{X}_s$, the definition of hop encoding is:

$$\mathbf{X}_h = [\underbrace{\mathbf{E}_h(0), \overbrace{\mathbf{E}_h(1), \mathbf{E}_h(1), \mathbf{E}_h(1)}^{\text{1st hop}}, \ldots, \overbrace{\mathbf{E}_h(K), \mathbf{E}_h(K), \mathbf{E}_h(K)}^{K\text{-th hop}},}_{\text{1st relation}} \quad (4)$$
$$\ldots, \underbrace{\mathbf{E}_h(0), \mathbf{E}_h(1), \mathbf{E}_h(1), \mathbf{E}_h(1), \ldots, \mathbf{E}_h(K), \mathbf{E}_h(K), \mathbf{E}_h(K)}_{R\text{-th relation}}]$$

where $\mathbf{E}_h(\cdot)$ denotes learnable embeddings that stored as a $(K+1) \times d_H$ matrix. As shown in Fig. 2, the boxed number (index) $k$ represents the corresponding one-hot vector and thus, the group vector of the $k$-th hop in the input sequence will be assigned a learnable embedding $\mathbf{E}_h(1)$ (e.g., when $K = 2$ and $k = 1$, $\mathbf{E}_h(1) = [0, 1, 0] \times W_{(K+1) \times d_H}$).

*4.3.3 Relation encoding.* Considering the complexity and diversity of the relational information between entities, distinguishing contributive relations is vital for fraud detection task. Here, we directly add a relation encoding $X_r$ to $X_s$:

$$X_r = [\underbrace{E_r(0), E_r(0), \ldots, E_r(0)}_{\text{1st relation}}, \underbrace{E_r(1), E_r(1), \ldots, E_r(1)}_{\text{2nd relation}}, \ldots, \\ \underbrace{E_r(R-1), E_r(R-1), \ldots, E_r(R-1)]}_{R-\text{th relation}} \quad (5)$$

where $E_r(\cdot)$ denotes learnable embeddings that stored as a $R \times d_H$ matrix. The Relation Encoding explicitly indicates the types of relations and assists the pair-wise self-attention to re-weight the group vectors from different relations.

*4.3.4 Group encoding.* In real scenarios, the labeled data accumulates rapidly. Most prior approaches learn representations mainly on entity features and only use labels as supervision signals at the training time. Unlike these methods, we exploit the partially observed entity labels to augment the original feature space. Inspired by the label encoding strategy in UniMP [22] that projects the label to the feature dimension, GAGA adds a learnable embedding to the input sequence $X_s$. In the group aggregation step, the neighbors are grouped according to the class labels, so the group embedding indexes are the same as the class labels. The group encoding $X_g$ is:

$$X_g = [\underbrace{\overbrace{E_g(*), E_g(-), E_g(+)}^{\text{1st hop}}, E_g(*), \ldots, \overbrace{E_g(-), E_g(+), E_g(*)}^{K-\text{th hop}}}_{\text{1st relation}}, \\ \ldots, \underbrace{E_g(*), E_g(-), E_g(+), E_g(*), \ldots, E_g(-), E_g(+), E_g(*)]}_{R-\text{th relation}} \quad (6)$$

where $E_g(\cdot)$ denotes learnable embeddings that stored as a $P \times d_H$ matrix. Note that the target node is encoded with $E_g(*)$ to avoid label leakage.

*4.3.5 Transformer encoder.* After computing the three learnable encodings for each group vector (the size of $X_h$, $X_r$, $X_g$ is $S$, the same as $X_s$), we fuse them and finally the input feature sequence of the transformer encoder is:

$$X_{in} = X_s + X_h + X_r + X_g. \quad (7)$$

In a transformer encoder, the multi-head attention modules are used to perform deep interaction among vectors. We update the hidden vector $x_i$ of the $i$-th group vector in $X_{in}$ from layer $l$ to $l + 1$:

$$\begin{aligned}
x_i^0 &= x_i, \\
x_i^{l+1} &= \text{Concat}(\text{head}_1, \ldots, \text{head}_M) O^l, \\
\text{head}_m &= \text{Attention}\left(Q^{m,l} x_i^l, K^{m,l} x_j^l, V^{m,l} x_j^l\right) \\
&= \sum_{j \in S} w_{ij}\left(V^{m,l} x_j^l\right), \\
w_{ij} &= \text{softmax}_j \left(\frac{Q^{m,l} x_i^l \cdot K^{m,l} x_j^l}{\sqrt{d_H}}\right),
\end{aligned} \quad (8)$$

where $Q^{m,l}$, $K^{m,l}$, $V^{m,l}$ are the learnable weights of the $m$-th attention head, $x_j^l$ is the $j$-th hidden vector of layer $l$ and $O^l$ is a projection to match the dimensions between adjacent layers. Additionally, we use Layer Normalization [1] and a position-wise 2-layer MLP after the multi-head attention module.

Table 1: Statistics of Datasets. IR represents the class imbalance ratio. $\varphi_r = \frac{|\{(u,v)|A_r[u,v]=1 \wedge y_u=y_v\}|}{\sum A_r[u,v]}$ is the homophily ratio which calculates the proportion of the immediate neighbors that share the same class label.

| Dataset | #Nodes (IR) | Relations | #Relations | $\varphi_r$ | #Feat |
|---|---|---|---|---|---|
| Amazon | 11,944 (13.5) | U-P-U | 175,608 | 0.1673 | 25 |
|  |  | U-S-U | 3,566,479 | 0.0558 |  |
|  |  | U-V-U | 1,036,737 | 0.0532 |  |
| YelpChi | 45,954 (5.9) | R-U-R | 49,315 | 0.9089 | 32 |
|  |  | R-S-R | 3,402,743 | 0.1857 |  |
|  |  | R-T-R | 573,616 | 0.1764 |  |
| BF10M | 13,251,571 (12.4) | U-J-U | 5,815,738 | 0.4092 | 161 |
|  |  | U-L-U | 65,530,647 | 0.1371 |  |

## 4.4 Inter-Aggregation for Prediction

In order to make predictions and aggregate cross-relation representations, we introduce inter-aggregation layer (Fig. 2 ❺). The output sequence of the last layer of the encoder is $Z$. We extract the corresponding output vectors $\{z_r | \forall r \in R\}$ in $Z$ and then aggregate them into the final representation $z_v$. Concretely, $z_r$ is the first output vector under relation $r$, so the indexes should be $\{0, s, 2 \times s, \ldots, (R-1) \times s\}$, where $s = P \times K + 1$. We can simply adopt concatenation operation as our candidate aggregator. Following the encoding and representation learning step, we append a MLP classifier as a new layer after the transformer encoder. The optimization objective of GAGA is defined as:

$$\mathcal{L} = -\sum_{v \in \hat{V}} [y_v \log p_v + (1 - y_v) \log(1 - p_v)] + \lambda \|\theta\|_2^2, \\ p_v = \text{sigmoid}(\text{MLP}(z_v)) \quad (9)$$

where $\theta$ is the parameter set of GAGA, $\lambda$ is the regularizer parameter.

## 5 EXPERIMENTS

The objective of the experimental evaluation is to investigate the overall performance of the proposed GAGA, as well as the effectiveness of the key components. The GAGA code[1] is available at *https://github.com/Orion-wyc/GAGA*.

### 5.1 Datasets

We first conduct experiments on two public spam review datasets, i.e. YelpChi and Amazon [4], each of which includes three different relations between entities. Moreover, we use a real-world industrial dataset collected by Baidu, namely BF10M. BF10M contains two relations: U-J-U connects websites if one visits another by a click triggered event; U-L-U connects two websites that link to each other. The target is to find the websites that involve in different forms of fraudulent activities, such as content plagiarism, spamming and phishing URLs. The statistics of the datasets are summarized in Tab. 1. Detailed description of the two public datasets can be found in Appendix.

---
[1] Archived DOI: https://doi.org/10.5281/zenodo.7608986

Table 2: Performance Comparison on public spam review datasets.

| Methods | YelpChi | | | Amazon | | |
| --- | --- | --- | --- | --- | --- | --- |
| | AUC | AP | F1-macro | AUC | AP | F1-macro |
| GCN$_{(ICLR'17)}$ | 0.5924±0.0030 | 0.2176±0.0119 | 0.5072±0.0271 | 0.8405±0.0075 | 0.4660±0.0131 | 0.6985±0.0046 |
| GAT$_{(ICLR'17)}$ | 0.6796±0.0070 | 0.2807±0.0048 | 0.5773±0.0080 | 0.8096±0.0113 | 0.3082±0.0067 | 0.6681±0.0076 |
| HAN$_{(WWW'19)}$ | 0.7420±0.0009 | 0.2722±0.0036 | 0.5472±0.0097 | 0.8421±0.0062 | 0.4631±0.0185 | 0.7016±0.0126 |
| GraphSAGE$_{(NeurIPS'17)}$ | 0.7409±0.0000 | 0.3258±0.0000 | 0.6001±0.0002 | 0.9172±0.0001 | 0.8268±0.0002 | 0.9029±0.0004 |
| Cluster-GCN$_{(KDD'19)}$ | 0.7623±0.0069 | 0.3691±0.0179 | 0.6204±0.0557 | 0.9211±0.0256 | 0.8075±0.0566 | 0.8853±0.0272 |
| GraphSAINT$_{(ICLR'20)}$ | 0.7412±0.0143 | 0.3641±0.0304 | 0.5974±0.0728 | 0.8946±0.0176 | 0.7956±0.0091 | 0.8888±0.0244 |
| CARE-GNN$_{(CIKM'20)}$ | 0.7854±0.0111 | 0.3972±0.0208 | 0.6064±0.0186 | 0.8823±0.0305 | 0.7609±0.0904 | 0.8592±0.0574 |
| FRAUDRE$_{(ICDM'21)}$ | 0.7588±0.0078 | 0.3870±0.0186 | 0.6421±0.0135 | 0.9308±0.0180 | 0.8433±0.0089 | 0.9037±0.0031 |
| PC-GNN$_{(WWW'21)}$ | 0.8154±0.0031 | 0.4797±0.0064 | 0.6523±0.0197 | 0.9489±0.0067 | 0.8435±0.0166 | 0.8897±0.0144 |
| RioGNN$_{(TOIS'21)}$ | 0.8144±0.0050 | 0.4722±0.0079 | 0.6422±0.0233 | 0.9558±0.0019 | 0.8700±0.0044 | 0.8848±0.0125 |
| H$^2$-FDetector$_{(WWW'22)}$ | 0.8892±0.0020 | 0.5543±0.0135 | 0.7345±0.0086 | 0.9605±0.0008 | 0.8494±0.0023 | 0.8010±0.0058 |
| SIGN$_{(ICML-GRL'20)}$ | 0.8569±0.0051 | 0.5801±0.0191 | 0.7308±0.0053 | 0.9404±0.0033 | 0.8483±0.0031 | 0.9046±0.0012 |
| GA+RNN$_{(Ablation)}$ | 0.9073±0.0237 | 0.6727±0.0672 | 0.7713±0.0313 | 0.9563±0.0075 | 0.8688±0.0086 | 0.9081±0.0055 |
| GA+LSTM$_{(Ablation)}$ | 0.9278±0.0025 | 0.7358±0.0119 | 0.7994±0.0041 | 0.9539±0.0089 | 0.8655±0.0117 | 0.9079±0.0066 |
| GT$_{(Ablation)}$ | 0.9084±0.0065 | 0.6979±0.0143 | 0.7839±0.0091 | 0.9514±0.0078 | 0.8581±0.0085 | **0.9137±0.0035** |
| GAGA$_{(Ours)}$ | **0.9439±0.0016** | **0.8014±0.0063** | **0.8323±0.0041** | **0.9629±0.0052** | **0.8815±0.0095** | 0.9133±0.0040 |

Table 3: Performance Comparison on BF10M.

| Methods | BF10M | | |
| --- | --- | --- | --- |
| | AUC | AP | F1-macro |
| MLP | 0.9066±0.0015 | 0.5135±0.0063 | 0.6840±0.1073 |
| GCN | 0.9194±0.0034 | 0.5282±0.0128 | 0.7478±0.0037 |
| GAT | 0.9143±0.0336 | 0.5487±0.0855 | 0.7469±0.0340 |
| GraphSAGE | 0.9597±0.0020 | 0.6965±0.0107 | 0.8103±0.0062 |
| Cluster-GCN | 0.9480±0.0039 | 0.6383±0.0221 | 0.7898±0.0084 |
| GraphSAINT | 0.9659±0.0016 | 0.7156±0.0092 | 0.8227±0.0061 |
| SIGN | 0.9652±0.0112 | 0.7262±0.0592 | 0.8228±0.0225 |
| H$^2$-FDetector | | OOM | |
| GAGA$_{(Ours)}$ | **0.9923±0.0004** | **0.9249±0.0036** | **0.9097±0.0032** |

## 5.2 Compared Methods

We select representative methods for overall comparison:

- **GCN** [10], **GAT** [25], **HAN** [26] represent general GNN models for homogeneous or heterogeneous graphs.
- **GraphSAGE** [6], **Cluster-GCN** [3], **GraphSAINT** [31] represent sampling-based GNN models.
- **CARE-GNN** [4], **RioGNN** [19], **PC-GNN** [14] are the previous state-of-the-art GNN-based fraud detectors with neighbor filtering modules. **FRAUDRE** [32] copes with the graph inconsistencies and class imbalance. **H$^2$-FDetector** [21] is a fraud-detector considers the effect of homophilic and heterophilic connections.
- **SIGN** [20] is a scalable architecture that use linear propagation layers to capture high-order information.
- **GAGA** is the proposed model. **GT (i.e., GAGA w/o GA)** is the variant of GAGA that removes group aggregation and learnable encodings. **GA+RNN** and **GA+LSTM** are variants that integrate group aggregation and other networks.
- **LPA** [35], **C&S** [7], **BoT** [22, 28] are selected to compare the performance improvement with the class labels.

## 5.3 Evaluation Metrics

We select three widely-used and complementary metrics:

- **AUC**: The Area Under Receiver Operating Characteristic, a ranking related metric that have no bias to any class for class imbalance classification in fraud detection.
- **AP**: The Area Under the Precision Recall Curve, which pays more attention to the ranking of fraudulent entities than that of benign entities [18].
- **F1-score**: **F1-fraud** for the positive class, **F1-benign** for the negative class and **F1-macro** for both.

The value of the above metrics are within [0, 1], and a larger one implies better performance. Ten-run average value and standard deviation on the testing set is reported.

## 5.4 Experimental Setting

In our experiments, Adam is selected as the optimizer, the patience of early-stopping is set to 100. The size of training/validation/testing set of the datasets is set to 0.4/0.1/0.5 for all the compared methods. For CARE-GNN, RioGNN, FRAUDRE and PC-GNN, H$^2$-FDetector we use the source code provided by the authors. GAGA provides dual implementations in both PaddlePaddle and PyTorch. The details on experimental settings are provided in Appendix.

## 5.5 Overall Performance Evaluation

As shown in in Tab. 2, GAGA outperforms all other competitive GNN methods on both datasets and achieves improvement on YelpChi by up-to 24.39% (compared with FRAUDRE). Since pure

Table 4: Comparison of label utilization methods.

| Methods | AUC | AP | F1-fraud | F1-benign |
|---|---|---|---|---|
| LPA | 0.6248 | 0.2567 | 0.2903 | 0.7021 |
| GCN | 0.5894 | 0.2032 | 0.2685 | 0.7085 |
| GCN (w/ C&S) | 0.6188↑ | 0.2588↑ | 0.2864↑ | 0.7301↑ |
| GAT | 0.6707 | 0.2737 | 0.3375 | 0.7872 |
| GAT (w/ C&S) | 0.6762↑ | 0.2875↑ | 0.3414↑ | 0.8083↑ |
| GT | 0.9059 | 0.6802 | 0.6322 | 0.9308 |
| GT (w/ C&S) | 0.9241↑ | 0.7352↑ | 0.6596↑ | 0.9390↑ |
| GT (w/ BoT) | 0.9281↑ | 0.7478↑ | 0.6672↑ | 0.9352↑ |
| GAGA (Ours) | **0.9423** | **0.7944** | **0.7135** | **0.9525** |

GNNs face *Challenge 1*, the result proves that the low homophily problem could hamper the performance of GNN-based fraud detectors. Although the homophily ratio of the fraudulent entities is low, GAGA still achieves better AP scores, which shows that the detector accurately returns a majority of all fraudsters. Among all the compared methods, CARE-GNN, FRAUDRE, PC-GNN and RioGNN address camouflaged fraudsters by filtering neighbors. We observe that the enhanced GNN-based detectors do show some improvements over those classic GNN models. H$^2$-FDetector is better than other GNN-based methods which indicates that tackling the low homophily issue can find more fraudulent entities. Besides, the AUC of SIGN is close to that of PC-GNN on Amazon and even better on YelpChi, which shows the worth of multi-hop neighborhood information. As shown in Tab. 3, GAGA also significantly outperforms other methods in industrial web anti-fraud tasks. We do not evaluate CARE-GNN, PC-GNN, FRAUDRE and Rio-GNN because the implementation of these methods fail to run.

### 5.6 Label Utilization Comparison

The main advantage of GAGA is utilizing the label information for feature augmentation. The aforementioned learning methods, including LPA, C&S, and BoT/UniMP, have been proven to be simple and effective tricks for label utilization. Tab. 4 compares different label utilization methods. We evaluate these methods on YelpChi and report single-run results because of the post-processing nature of C&S. Note also that two rows without a split line are based on the same trained model (e.g., GCN and GCN (w C&S)). From Tab. 4, we can observe that GAGA achieves the best performance and significantly outperforms other compared methods in all metrics. LPA, C&S and BoT are powerful baselines in common tasks, but they have worse performance in the low homophily setting. For GCN, GAT and GT, the performance of C&S + GNN exceeds that of vanilla GNNs without label utilization by 0.59~4.99% (AUC), 1.14~27.36% (AP), 1.16~6.67% (F1-fraud). The most interesting finding is that LPA outperforms GCN with 6.01% in AUC, which indicates that overlooking the observed labels during training and inference is a waste of information.

### 5.7 Ablation Study and Parameter Analysis

*5.7.1 The effectiveness of each component of GAGA.* To answer how each component contributes to the performance of GAGA, we replace parts of the entire framework for ablation study. The results are reported in Tab. 2 (last four rows). First, when we replace the GA with simple mean aggregator and keep the backbone encoder the same, the performance drops by 3.8%, 12.9% and 5.8% in terms of AUC, AP and F1-macro on YelpChi. This demonstrates that GA can mitigate the low homophily issue. And secondly, we combine GA with other backbone encoders. We can observe that the performance drop by 1.8% in AUC, 6.9% in AP and 3.1% in F1-macro across public datasets. We can tell from the results that Transformer encoder can better capture the semantic information of neighborhood, indicating that multi-head self-attention can adaptively re-weight the group vectors of different property. Furthermore, the result proves that GA is effective and portable: it can be conveniently equipped to other backbone networks for improving performance in the low homophily setting.

Table 5: Ablation study to demonstrate the effectiveness of the Learnable Encodings.

| Ablation (n_hops=2) | | AUC | |
|---|---|---|---|
| | | YelpChi | Amazon |
| w/ GA | $X_g, X_r, X_h$ | **0.9439±0.0016** | **0.9629±0.0052** |
| | $X_r, X_h$ | 0.9274±0.0028 | 0.9584±0.0082 |
| | $X_g$ | 0.9355±0.0032 | 0.9584±0.0099 |
| | $X_r$ | 0.9217±0.0036 | 0.9581±0.0106 |
| | $X_h$ | 0.9144±0.0039 | 0.9540±0.0099 |
| | – | 0.9020±0.0066 | 0.9524±0.0070 |
| w/o GA | $X_r, X_h$ | 0.9219±0.0047 | 0.9578±0.0095 |
| | $X_r$ | 0.9197±0.0048 | 0.9574±0.0075 |
| | $X_h$ | 0.9172±0.0046 | 0.9520±0.0081 |
| | – | 0.9084±0.0065 | 0.9528±0.0077 |

Table 6: Sensitivity analysis to verify the performance under different percentage (1~40%) of training data and percentage (1~40%) of partially observed labels. (YelpChi)

| Train (%) | Label (%) | AUC | AP |
|---|---|---|---|
| 40 | 40 | **0.9439±0.0016** | **0.8014±0.0063** |
| 40 | 30 | 0.9416±0.0020 | 0.7904±0.0067 |
| 40 | 20 | 0.9328±0.0034 | 0.7619±0.0112 |
| 40 | 10 | 0.9232±0.0024 | 0.7335±0.0060 |
| 30 | 30 | 0.9320±0.0030 | 0.7641±0.0076 |
| 20 | 20 | 0.9081±0.0189 | 0.6894±0.0456 |
| 10 | 10 | 0.8782±0.0107 | 0.5924±0.0207 |
| 5 | 5 | 0.8130±0.0064 | 0.4703±0.0315 |
| 1 | 1 | 0.7000±0.0710 | 0.3104±0.0744 |

*5.7.2 The effect of the learnable encodings.* To verify the effectiveness of the learnable encodings, we compare different compositions of the encodings. In Tab. 5, GAGA achieves the best performance with complete learnable encodings. The group encoding $X_g$ solely contributes 3.71% improvement in AUC. The variant without group aggregation performs worse in most cases, which demonstrates the effectiveness of the group aggregation and group encoding.

*5.7.3 The effect of training rate and label rate.* Tab. 6 illustrates the sensitivity of the training percentage and the label percentage on YelpChi. Varying the percentage of training data from 10% to 40%, there is a 7.48% increase in AUC. Even though only 10% of the nodes are used, GAGA outperforms the GNN-based detectors (with 40% training data in Tab. 2). This demonstrates that if the labeled neighbors only account for a small portion, GAGA can still achieve performance improvement. We also report the performance of GAGA when the labeled nodes are extremely rare (the lower half, 1% and 5% training nodes). GAGA still shows a competitive result. Besides, we explore the effect of the percentage of the ground truth labels in training and prediction phase in Tab. 6 (the upper half). It is observed that the accumulated ground-truth labels can boost the performance of fraud detection.

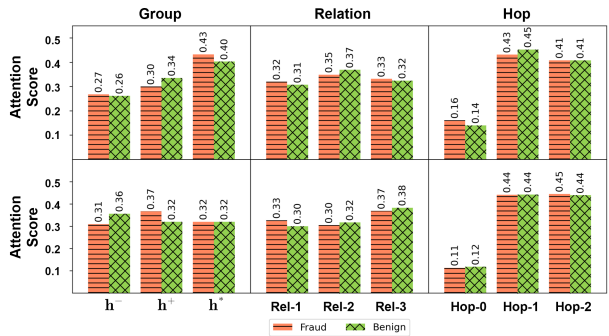

Figure 3: Visualization of attention scores on Amazon (upper) and YelpChi (lower) based on statistics. (Rel-1, Rel-2, Rel-3) denotes (U-P-U, U-S-U, U-V-U) for Amazon and (R-S-R, R-T-R, R-U-R) for YelpChi, respectively.

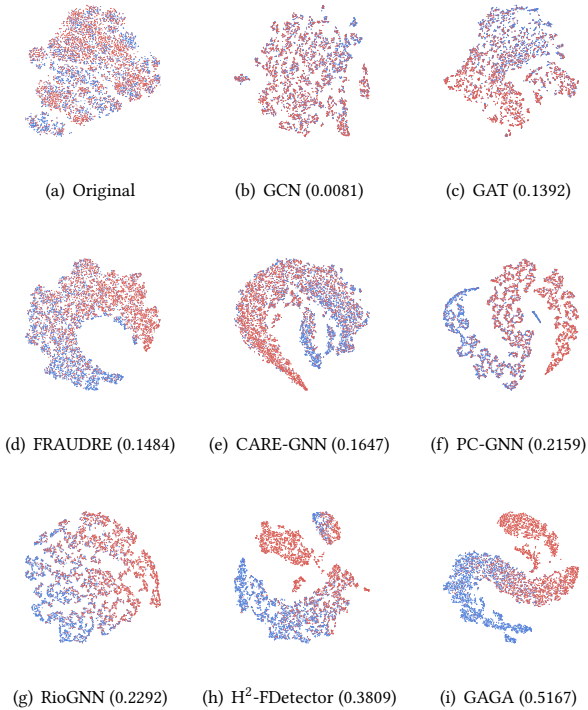

Figure 4: The t-SNE visualization of latent representations and the Adjusted Rand Index of each cluster. (Dataset: YelpChi, Fraud: red, Benign: blue)

Table 7: Comparison of training throughput (sample/s).

| Methods | YelpChi | Amazon | BF10M |
|---|---|---|---|
| GAGA | 21739.13 | 17241.38 | 12820.51 |
| FRAUDRE | 244.92 | 403.06 | - |
| CARE-GNN | 13888.89 | 3125.00 | - |
| PC-GNN | 9900.99 | 1937.98 | - |
| RioGNN | 17241.38 | 3436.43 | - |
| $H^2$-FDetetor | 11185.72 | 336.60 | OOM |

### 5.8 Visualization

To intuitively demonstrate the effectiveness of GAGA, the attention scores of positive and negative entities are visualized in Fig. 3. Column 1 presents that fraud entities tends to retain larger attention scores on group $\mathbf{h}^+$ than that on $\mathbf{h}^-$. Column 2 shows that GAGA could adaptively select different relations. For YelpChi, both classes get a larger score on relation R-U-R since this relation has a extremely higher homophily ratio (0.9089). Column 3 is the attention score among hops. Different from CARE-GNN and PCGNN that report performance decline as the number of layers increases, GAGA achieves a further improvement with high-order neighborhood information. Moreover, We visualize the node embeddings of different methods in Fig. 4. It is clearly observed that GAGA generates notably better embeddings that distinguishes the fraudulent entities from the benign ones. To verify our conclusion, we also perform the K-Means clustering ($K = 2$) on the 2-dimensional embeddings after t-SNE and report the Adjusted Rand Index (ARI) of each cluster (higher is better).

### 5.9 End-to-End Performance

In this experiment, we compare the end-to-end training performance among GNN-based fraud detectors across all datasets. We observe that GAGA obtains the highest throughput during training, which is critical for processing large-scale multi-relational graphs. GAGA decouples the message passing and the representation learning, so the training throughput is independent of the structure of the original graph. In contrast, the training throughput of the five GNN-based approaches drops when the average degree of the graph is high. Time complexity analysis is provided in Appendix.

## 6 CONCLUSION AND FUTURE WORK

In this paper, we present a transformer-based method for fraud detection in multi-relation graphs. Enhanced by the group aggregation and the learnable encodings, GAGA solves both challenges in graph-based fraud detection. Experimental results on both public and real-world industrial dataset demonstrate the effectiveness of GAGA. For future work, generalizing GAGA to multi-class node classification could be a promising direction.

Label Information Enhanced Fraud Detection against Low Homophily in Graphs    Conference '23, April 30 – May 4,2023, Austin, Texas, USA## REFERENCES
[1] Lei Jimmy Ba, Jamie Ryan Kiros, and Geoffrey E. Hinton. 2016. Layer normalization. arXiv:1607.06450
[2] Deng Cai and Wai Lam. 2020. Graph transformer for graph-to-sequence learning. In *Proceedings of the 34th AAAI Conference on Artificial Intelligence*, Vol. 34. 7464–7471.
[3] Wei-Lin Chiang, Xuanqing Liu, Si Si, Yang Li, Samy Bengio, and Cho-Jui Hsieh. 2019. Cluster-gcn: an efficient algorithm for training deep and large graph convolutional networks. In *Proceedings of the 25th ACM SIGKDD International Conference on Knowledge Discovery & Data Mining*. 257–266.
[4] Yingtong Dou, Zhiwei Liu, Li Sun, Yutong Deng, Hao Peng, and Philip S Yu. 2020. Enhancing graph neural network-based fraud detectors against camouflaged fraudsters. In *Proceedings of the 29th ACM International Conference on Information & Knowledge Management*. 315–324.
[5] Yingtong Dou, Guixiang Ma, Philip S Yu, and Sihong Xie. 2020. Robust spammer detection by nash reinforcement learning. In *Proceedings of the 26th ACM SIGKDD International Conference on Knowledge Discovery & Data Mining*. 924–933.
[6] Will Hamilton, Zhitao Ying, and Jure Leskovec. 2017. Inductive representation learning on large graphs. In *Advances in Neural Information Processing Systems*, Vol. 30.
[7] Qian Huang, Horace He, Abhay Singh, Ser-Nam Lim, and Austin R Benson. 2020. Combining label propagation and simple models out-performs graph neural networks. arXiv:2010.13993
[8] Yugang Ji, Guanyi Chu, Xiao Wang, Chuan Shi, Jianan Zhao, and Junping Du. 2022. Prohibited item detection via risk graph structure learning. In *Proceedings of the ACM Web Conference 2022*. 1434–1443.
[9] Parisa Kaghazgaran, Majid Alfifi, and James Caverlee. 2019. Wide-ranging review manipulation attacks: model, empirical study, and countermeasures. In *Proceedings of the 28th ACM International Conference on Information and Knowledge Management*. 981–990.
[10] Thomas N. Kipf and Max Welling. 2017. Semi-supervised classification with graph convolutional networks. In *Proceedings of the 5th International Conference on Learning Representations*.
[11] Ao Li, Zhou Qin, Runshi Liu, Yiqun Yang, and Dong Li. 2019. Spam review detection with graph convolutional networks. In *Proceedings of the 28th ACM International Conference on Information and Knowledge Management*. 2703–2711.
[12] Guohao Li, Chenxin Xiong, Ali Thabet, and Bernard Ghanem. 2020. Deepergcn: all you need to train deeper gcns. arXiv:2006.0773
[13] Xue Li and Yuanzhi Cheng. 2022. Irregular message passing networks. *Knowledge-Based Systems* (2022), 109919. https://doi.org/10.1016/j.knosys.2022.109919
[14] Yang Liu, Xiang Ao, Zidi Qin, Jianfeng Chi, Jinghua Feng, Hao Yang, and Qing He. 2021. Pick and choose: a GNN-based imbalanced learning approach for fraud detection. In *Proceedings of the Web Conference 2021*. 3168–3177.
[15] Yixin Liu, Shirui Pan, Yu Guang Wang, Fei Xiong, Liang Wang, Qingfeng Chen, and Vincent CS Lee. 2021. Anomaly detection in dynamic graphs via transformer. *IEEE Transactions on Knowledge and Data Engineering* (2021).
[16] Zhiwei Liu, Yingtong Dou, Philip S Yu, Yutong Deng, and Hao Peng. 2020. Alleviating the inconsistency problem of applying graph neural network to fraud detection. In *Proceedings of the 43rd international ACM SIGIR Conference on Research and Development in Information Retrieval*. 1569–1572.
[17] Xiaoxiao Ma, Jia Wu, Shan Xue, Jian Yang, Chuan Zhou, Quan Z Sheng, Hui Xiong, and Leman Akoglu. 2021. A comprehensive survey on graph anomaly detection with deep learning. *IEEE Transactions on Knowledge and Data Engineering* (2021).
[18] Christopher Manning, Prabhakar Raghavan, and Hinrich Schütze. 2010. Introduction to information retrieval. *Natural Language Engineering* 16, 1 (2010), 100–103.
[19] Hao Peng, Ruitong Zhang, Yingtong Dou, Renyu Yang, Jingyi Zhang, and Philip S Yu. 2021. Reinforced neighborhood selection guided multi-relational graph neural networks. *ACM Transactions on Information Systems* 40, 4 (2021), 1–46.
[20] Emanuele Rossi, Fabrizio Frasca, Ben Chamberlain, Davide Eynard, Michael M. Bronstein, and Federico Monti. 2020. SIGN: scalable inception graph neural networks. arXiv:2004.11198
[21] Fengzhao Shi, Yanan Cao, Yanmin Shang, Yuchen Zhou, Chuan Zhou, and Jia Wu. 2022. H2-FDetector: a gnn-based fraud detector with homophilic and heterophilic connections. In *Proceedings of the ACM Web Conference 2022*. 1486–1494.
[22] Yunsheng Shi, Zhengjie Huang, Shikun Feng, Hui Zhong, Wenjing Wang, and Yu Sun. 2021. Masked label prediction: unified message passing model for semi-supervised classification. In *Proceedings of the 30th International Joint Conference on Artificial Intelligence*. 1548–1554.
[23] Yu Sun, Shuohuan Wang, Yukun Li, Shikun Feng, Xuyi Chen, Han Zhang, Xin Tian, Danxiang Zhu, Hao Tian, and Hua Wu. 2019. Ernie: enhanced representation through knowledge integration. arXiv:1904.09223
[24] Ashish Vaswani, Noam Shazeer, Niki Parmar, Jakob Uszkoreit, Llion Jones, Aidan N Gomez, Łukasz Kaiser, and Illia Polosukhin. 2017. Attention is all you need. In *Advances in Neural Information Processing Systems*, Vol. 30.
[25] Petar Veličković, Guillem Cucurull, Arantxa Casanova, Adriana Romero, Pietro Lio, and Yoshua Bengio. 2017. Graph attention networks. In *Proceedings of the 5th International Conference on Learning Representations*.
[26] Xiao Wang, Houye Ji, Chuan Shi, Bai Wang, Yanfang Ye, Peng Cui, and Philip S Yu. 2019. Heterogeneous graph attention network. In *Proceedings of the ACM Web Conference 2019*. 2022–2032.
[27] Yangkun Wang, Jiarui Jin, Weinan Zhang, Yongyi Yang, Jiuhai Chen, Quan Gan, Yong Yu, Zheng Zhang, Zengfeng Huang, and David Wipf. 2022. Why propagate alone? parallel use of labels and features on graphs. In *The 10th International Conference on Learning Representations*.
[28] Yangkun Wang, Jiarui Jin, Weinan Zhang, Yong Yu, Zheng Zhang, and David Wipf. 2021. Bag of Tricks for Node Classification with Graph Neural Networks.
[29] Zonghan Wu, Shirui Pan, Fengwen Chen, Guodong Long, Chengqi Zhang, and S Yu Philip. 2020. A comprehensive survey on graph neural networks. *IEEE Transactions on Neural Networks and Learning Systems* 32, 1 (2020), 4–24.
[30] Dong Yuan, Yuanli Miao, Neil Zhenqiang Gong, Zheng Yang, Qi Li, Dawn Song, Qian Wang, and Xiao Liang. 2019. Detecting fake accounts in online social networks at the time of registrations. In *Proceedings of the 2019 ACM SIGSAC Conference on Computer and Communications Security*. 1423–1438.
[31] Hanqing Zeng, Hongkuan Zhou, Ajitesh Srivastava, Rajgopal Kannan, and Viktor Prasanna. 2020. GraphSAINT: graph sampling based inductive learning method. *Proceedings of the 8th International Conference on Learning Representations* (2020).
[32] Ge Zhang, Jia Wu, Jian Yang, Amin Beheshti, Shan Xue, Chuan Zhou, and Quan Z Sheng. 2021. FRAUDRE: fraud detection dual-resistant to graph inconsistency and imbalance. In *Proceedings of the 2021 IEEE International Conference on Data Mining*. 867–876.
[33] Mingxue Zhang, Wei Meng, Sangho Lee, Byoungyoung Lee, and Xinyu Xing. 2019. All your clicks belong to me: investigating click interception on the web. In *Proceedings of the 28th USENIX Security Symposium*. 941–957.
[34] Jiong Zhu, Yujun Yan, Lingxiao Zhao, Mark Heimann, Leman Akoglu, and Danai Koutra. 2020. Beyond homophily in graph neural networks: Current limitations and effective designs. In *Advances in Neural Information Processing Systems*, Vol. 33. 7793–7804.
[35] Xiaojin Zhu, Zoubin Ghahramani, and John D Lafferty. 2003. Semi-supervised learning using gaussian fields and harmonic functions. In *Proceedings of the 20th International conference on Machine learning*. 912–919.
## A APPENDIX

### A.1 From Message Passing to Group Aggregation

The complete procedure of group aggregation is summarized in Algorithm 1. In this section, we will explain the rationality of the group aggregation from the perspective of general message passing. The group aggregation can be formulated as multiple separate message passing processes, which means the node representations are generated by aggregating the neighborhood in different groups. In this work, the group aggregation splits the neighbors based on the partially observed class labels.

With respect to a specific relation $r$, a class-limited group aggregation matrix $\mathbf{A}_{c,r}$ is derived by setting the same entries as the original adjacency matrix $\mathbf{A}_r$ while other entries are set to zero. $c$ is a specific class in the partially observed nodes $\hat{\mathcal{V}}$. The group aggregation matrix is defined as:

$$\mathbf{A}_{c,r}[u,v] = \begin{cases} \mathbf{A}_r[u,v], & v \in \hat{\mathcal{V}} \text{ and } y_v = c \\ 0, & \text{otherwise} \end{cases}. \quad (10)$$

For those unlabeled (masked) nodes (i.e, nodes in $\overline{\mathcal{V}} = \mathcal{V} - \hat{\mathcal{V}}$), the group aggregation matrix $\mathbf{A}_{*,r}$ is defined as:

$$\mathbf{A}_{*,r} = \begin{cases} \mathbf{A}_r[u,v], & v \in \overline{\mathcal{V}} \\ 0, & \text{otherwise} \end{cases} \quad (11)$$

Then, the aggregated neighborhood information of the immediate neighbors (1-hop) belonging to class $c$ is:

$$\mathbf{H}_{c,r} = \sigma(\mathbf{A}_{c,r}\mathbf{X}\mathbf{W}), \quad (12)$$

where $\mathbf{H}_{c,r}$ is the aggregated feature matrix, $\mathbf{X}$ is the original feature matrix and $\mathbf{W}$ is a learnable weight matrix. It is obvious that in the

**Algorithm 1:** Group Aggregation (GA)

**Input:** Target Node $v$, $K$ hop neighbors with self-loop $\{\hat{\mathcal{N}}_i(v)|i=1,2,\ldots,K\}$, partially observed nodes $\hat{\mathcal{V}}$ and labels $\hat{\mathcal{Y}}$, masked nodes $\overline{\mathcal{V}}$, number of classes $C$.

**Output:** Group vectors $\mathbf{H}_g = \{\mathbf{h}_1, \mathbf{h}_2, \ldots, \mathbf{h}_C, \mathbf{h}^*\}$

1 **Grouping($\hat{\mathcal{N}}_i(v)$):**
   // construct empty node groups
2  $V \leftarrow \{V_1, V_2, \ldots, V_C, V^*\}$
   // target node masking for $v$
3  Remove $\{v\}$ from $\hat{\mathcal{N}}_i(v)$
4  Put $\{v\}$ into group $V^*$
5  **for** $u \in \hat{\mathcal{N}}_i(v)$ **do**
6   **if** $u \in \hat{\mathcal{V}}$ **then**
      // Group id (class label)
7    gid = $y_u \in \hat{\mathcal{Y}}$
8    Put $\{u\}$ into group $V_{gid}$
9   **else if** $u \in \overline{\mathcal{V}}$ **then**
10   Put $\{u\}$ into group $V^*$
11  **return** $V$

12 **Aggregation():**
   // construct empty group vector
13  $\mathbf{H}_r \leftarrow []$
    // GA for each hop's neighbors
14  **for** $i \in 1, 2, \ldots, K$ **do**
15   $V \leftarrow$ Grouping($\hat{\mathcal{N}}_i(v)$)
16   $\mathbf{H}_g^{(k)} \leftarrow$ Aggregate $V$ via Equation (2)
17   Concatenate $\mathbf{H}_r$ and $\mathbf{H}_g^{(k)}$
18  **return** $\mathbf{H}_r$

1-hop setting, the $i$-th row vector of $\mathbf{H}_{c,r}$ is the same as $c$-th vector in $\mathbf{H}_g$ generated by Eq. 2. Note that the above analysis removes self-loops (the main diagonal elements are tagged as zero) since encoding the feature of the target node and its neighbors separately is proved to be effective in [34].

Considering the limitation of classic GNNs in aggregating high-order neighbors simultaneously, we perform a *shortcut* message passing with respect to the $k$-hop neighbors. The only difference is replacing the original adjacency matrix $\mathbf{A}_r$ with $\mathbf{A}_r^k$ (i.e., the $k$-th power of $\mathbf{A}_r$). Thus, we can get the high-order feature matrix $\mathbf{H}_{c,r}^{(k)}$ and each row vector represents a part in the input sequence $\mathbf{X}_s$.

### A.2 Computational Complexity

Here we discuss the computational complexity of GAGA (sinlge relation). Generally, group aggregation is rather efficient, including the message passing and aggregation. Group Aggregation is similar to GraphSAGE [6] but uses a flat aggregation. In this way, for every target node, we are guaranteed to have a bounded K-hop neighbourhood of $O(c^K)$ nodes, where $c$ is the average degree of the graph. Thus, the precise time complexity is $O(|V|c^K d)$ (for BF10M, $c_{U-J-U}$ = 1.88, $c_{U-L-U}$ = 10.89, and in our experiments, $K <= 4$). This is quite efficient in practical applications since the graph is processed only once and we can use data parallelism to accelerate it. With regard to the Transformer encoder, we consider a simple situation where $d_Q = d_K = d_V = d$ ($d_Q$, $d_K$ and $d_V$ represent the dimensions of Q, K, V, respectively). Thus, the complexity of the self-attention layer is $O(S^2 d + Sd^2)$. The overall time complexity of GAGA is $O(|V|c^K d + eL(S^2 d + Sd^2))$, where $e$ is the number of training epochs, $L$ is the number of attention layers.

Besides, most GNN-based fraud detectors face the high time complexity during training when the number of GNN layers increases. For example, GCN is trained using full-batch gradient descent, so the time complexity of a K-layer model is $O(eK|V|d^2)$. As for the minibatch training, we can get the time complexity of $O(e|V|Kc^K d^2)$. This reduces the memory consumption but introduces extra computation.

**Table 8: The hyper-paramerter setting of GAGA.**

| Parameter | Dataset | | |
|---|---|---|---|
| | YelpChi | Amazon | BF10M |
| learning rate | 0.001 | 0.001 | 0.001 |
| weight decay | 0.0001 | 0.0001 | 0.0001 |
| batchsize | 512 | 256 | 60000 |
| dropout | 0.1 | 0.0 | 0.1 |
| n_hidden | 32 | 16 | 64 |
| n_layers | 3 | 2 | 4 |
| n_head | 4 | 4 | 4 |
| n_hops | 2 | 2 | 1 |
| $\alpha$ | 1.0 | 1.0 | 1.0 |
| max epoch | 500 | 500 | 300 |

### A.3 Detailed Dataset Description

YelpChi and Amazon are opinion fraud detection dataset released in [4, 16]. The YelpChi dataset collects hotel and restaurant reviews on Yelp and the Amazon contains product reviews under the Musical Instrument category.

**YelpChi** takes reviews as the nodes and includes three relations: 1) Reviews posted by the same user are connected under **R-U-R**; 2)Reviews with the same stars of the same product are connected under **R-S-R**; 3)Reviews of the same product and posted in the same month are connected under **R-T-R**.

**Amazon** takes users as the nodes, which also contains three relations: 1) Users that review at least one same product are connected under **U-P-U**. 2) Users with at least one same stars within one week are connected under **U-S-U**. 3) Users with top-5% mutual review TF-IDF similarities are connected under **U-V-U**.

### A.4 Implementation Details

The proposed GAGA provides dual implementations in both PaddlePaddle and PyTorch. For BF10M, the group aggregation step is preprocessed by MapReduce and distributed data parallelism is launched to accelerate training if possible. All experiments are run on a server with 48 cores, 512GB memory and 8 NVIDIA Tesla V100 GPUs. The hyper-paramerter setting of GAGA is listed in Tab. 8. For CARE-GNN, PC-GNN, FRAUDRE, RioGNN and $H^2$-FDetector, we use the hyper-parameters provided in the original paper, and for the other baseline methods, we use unified configuration: embedding size (64 for YelpChi and Amazon, 128 for BF10M), learning rate (0.01), regularizer parameter $\lambda$ = 0.0001, batch size (512 for YelpChi, 256 for Amazon, 40960 for BF10M), number of layers (2). C&S is a post-processing method performed on trained models via hyperparameter optimization. We tune the following hyper-parameters: 1) the fraction of label information in correction step $\alpha_c = \{0.05, 0.1, 0.9, 0.95\}$; 2) the number of correction layers $n_c = \{0, 1, 2, 3\}$; 3) the fraction of label information in smoothing step $\alpha_s = \{0.05, 0.1, 0.9, 0.95\}$, 4) the number of smoothing layers $n_s = \{0, 1, 2, 3\}$. BoT/UniMP uses a masked label prediction method to supervised training the model. Here, the label_rates of masked labels are generated within the closed interval $[0.4, 0.625]$ of step size 0.025.